%% file: main.tex
\documentclass[10pt,twocolumn,letterpaper]{article}
\usepackage[pagenumbers]{iccv}     

\usepackage{url}
\usepackage{tabularx}
\usepackage{color-edits}
\usepackage{float}
\usepackage{upgreek}
\usepackage{amsbsy}
\usepackage{microtype}
\usepackage{amssymb}
\usepackage{amsmath}
\usepackage{amsthm}
\usepackage{graphicx}
\usepackage{setspace}
\usepackage{mathtools}
\usepackage[dvipsnames]{xcolor}
\usepackage{nicefrac}
\usepackage{enumitem}
\usepackage{caption}
\usepackage{adjustbox,booktabs,multirow}
\usepackage{autonum}

\definecolor{iccvblue}{rgb}{0.21,0.49,0.74}
\usepackage[pagebackref,breaklinks,colorlinks,allcolors=iccvblue]{hyperref}

\def\paperID{13085} 
\def\confName{ICCV}
\def\confYear{2025}

\input{math_commands.tex}

\definecolor{Gray}{gray}{0.9}

\author{\normalfont Vahid Balazadeh\thanks{This project was completed during an internship at Autodesk Research.}\\
University of Toronto\\
{\tt\small vahid@cs.toronto.edu}
\and
Mohammadmehdi Ataei\\
Autodesk Research\\
{\tt\small mehdi.ataei@autodesk.com}
\and
Hyunmin Cheong\\
Autodesk Research\\
{\tt\small hyunmin.cheong@autodesk.com}
\and\\
Amir Hosein Khasahmadi\\
Autodesk Research\\
{\tt\small amir.khasahmadi@autodesk.com}
\and\\
Rahul G. Krishnan\\
University of Toronto\\
{\tt\small rahulgk@cs.toronto.edu}
}

\title{Physics Context Builders: A Modular Framework for Physical Reasoning in Vision-Language Models}

\usepackage[accsupp]{axessibility}

\begin{document}

\maketitle

\begin{abstract}
Physical reasoning remains a significant challenge for Vision-Language Models (VLMs). This limitation arises from an inability to translate learned knowledge into predictions about physical behavior. Although continual fine-tuning can mitigate this issue, it is expensive for large models and impractical to perform repeatedly for every task. This necessitates the creation of modular and scalable ways to teach VLMs about physical reasoning. To that end, we introduce Physics Context Builders (PCBs), a modular framework where specialized smaller VLMs are fine-tuned to generate detailed physical scene descriptions. These can be used as physical contexts to enhance the reasoning capabilities of larger VLMs. PCBs enable the separation of visual perception from reasoning, allowing us to analyze their relative contributions to physical understanding. We perform experiments on CLEVRER and on Falling Tower, a stability detection dataset with both simulated and real-world scenes, to demonstrate that PCBs provide substantial performance improvements, increasing average accuracy by up to 13.8\% on complex physical reasoning tasks. Notably, PCBs also show strong Sim2Real transfer, successfully generalizing from simulated training data to real-world scenes.
\end{abstract}

\section{Introduction}

Physical reasoning is a fundamental component of human intelligence, enabling interpretation of complex interactions, prediction of future events, and understanding of causal relationships in real-world environments~\cite{kubricht2017intuitive}. For humans, the ability to understand the physical world is developed early and operates intuitively~\cite{mccloskey1983intuitive,baillargeon2004infants,teglas2011pure}. However, physical reasoning remains a significant challenge for artificial intelligence (AI) systems~\cite{lake2017building,srivastava2022beyond,yi2019clevrer,chow2025physbench}, despite advances in computer vision and Vision-Language Models (VLMs)~\cite{radford2021learning,alayrac2022flamingo,li2023blip,zhang2024vision,sun2024probing,li2024multimodal,ventura2025nleye}.

VLMs are remarkably successful on many predictive problems, spanning a broad range of tasks. This makes their use central to real-world applications requiring expertise in a variety of tasks, such as robotics and embodied AI~\cite{liu2024moka,guo2024phygrasp}. However, current VLMs consistently fail at physical reasoning tasks, struggling with basic spatial relationships (e.g., object positioning, counting)~\cite{rahmanzadehgervi2024vision}, object attributes~\cite{wang2023newton}, and physical interactions (e.g., stability assessment, dynamics prediction)~\cite{ghaffari2024exploring,chow2025physbench}. While humans leverage causal or physically guided knowledge for physical understanding of the world~\cite{battaglia2013simulation}, the mechanisms by which VLMs make predictions are not well understood. This leads to two important questions: \emph{what factors contribute to the lack of physical understanding in VLMs, and how can we improve it?}

One potential explanation for such limitations lies in VLMs' training data. VLMs are created by fusing the representations of image and text encoders using datasets like MSCOCO~\cite{lin2014microsoft} and Conceptual Captions~\cite{sharma2018conceptual}, which focus on general scene descriptions but lack annotations of physical relationships. Our experiments with standard benchmarks like CLEVRER~\cite{yi2019clevrer} confirm this hypothesis, demonstrating that fine-tuning on physics-focused data can enable a relatively small VLM to approach state-of-the-art results achieved by specialized architectures~\cite{yi2019clevrer,ding2021attention}. However, as the ecosystem of open and closed source models grows, repeatedly fine-tuning each VLM for every regime of physical concepts it cannot reason about becomes impractical; there is a need for practical, performant and modular tools to augment VLM capabilities.

\begin{figure*}[t]
     \centering
     \includegraphics[width=0.85\linewidth]{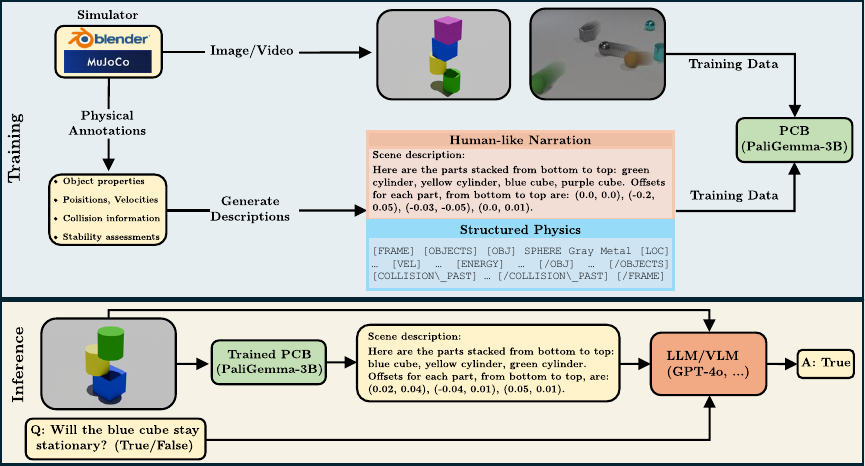}
    \caption{Physics Context Builders (PCBs) pipeline. The training phase (top) shows how physics simulators generate images/videos with corresponding annotations, which are converted into two types of physical descriptions: Human-like Narration (providing natural language scene descriptions with object properties and spatial relationships) and Structured Physics (offering frame-by-frame structured descriptions with physical properties). These descriptions serve as training data for fine-tuning a relatively small VLM into specialized PCBs. During the inference phase (bottom), a trained PCB processes a new image/video and generates detailed physical context about the scene, which is then provided to a foundation model (e.g., GPT-4o) alongside a user query to produce physically grounded responses.}
    \label{fig:data_gen_diagram}   
\end{figure*}

To this end, we introduce Physics Context Builders (PCBs): specialized VLMs that are fine-tuned on simulation data to generate fine-grained physical descriptions that can be used by larger VLMs. Simulation can provide precise annotations of controllable physical interactions of interest, which can serve as scalable training data to teach models about physical phenomena. We show how to embed the knowledge of physical concepts into PCBs and how they can transfer this knowledge to larger VLMs; this enables us to effectively separate visual perception from reasoning, using PCBs as perception modules, while leaving reasoning to larger models. See \cref{fig:data_gen_diagram} for a demonstration of PCBs.

PCBs offer a practical and modular approach to enhance the physical reasoning of existing large-scale VLMs without requiring expensive or infeasible modification of the larger foundation models. Our experiments on standard benchmarks demonstrate effective integration of PCBs with standard commercial VLMs. Moreover, they show that PCBs trained with simulation data successfully generalize to real-world scenarios, effectively performing simulation-to-reality (Sim2Real) transfer. In summary, our contributions are:
\begin{enumerate}[leftmargin=*]
    \item We introduce a modular framework where specialized VLMs are fine-tuned to generate detailed physical scene descriptions, enhancing the physical reasoning capabilities of foundation models without their modification.
    \item We demonstrate the effectiveness of separating visual perception from reasoning through our PCB approach, providing insights into how each component contributes to physical understanding.
    \item We show how simulation can be leveraged to train specialized modules that transfer successfully to real-world scenarios, avoiding expensive simulations at inference.
\end{enumerate}

\section{Related Work}

\subsection{Physical Reasoning in Vision Models}
Physical reasoning presents significant challenges for vision models, including both specialized architectures and large-scale VLMs. Recent benchmarks demonstrate that current VLMs struggle with basic physical understanding, often failing at tasks such as counting, depth reasoning, and physical interaction prediction \cite{ghaffari2024exploring,li2024multimodal,ventura2025nleye}. These limitations are not unique to VLMs; physical reasoning has also proven difficult for single-purpose models, necessitating specialized architectures such as physics-inspired predictive models \cite{battaglia2016interaction,guen2020disentangling,duan2022pip}, neural-symbolic executors \cite{chen2021grounding,ishay2024think,zheng2024contphy}, differentiable physics engines \cite{wu2015galileo,ding2021dynamic}, simulation-in-the-loop approaches \cite{liu2022mind,zhu2025maps}, and task-specific architectures \cite{ding2021attention}. However, large VLMs are not easily amenable to architectural modifications and require alternative approaches to enhance their capabilities, which we consider in this work.

\subsection{Evaluating Physical Reasoning in VLMs}
Several recent works have evaluated physical reasoning capabilities in vision-language models. \citet{nagar2024zero} benchmark zero-shot visual reasoning in both large language models (LLMs) and VLMs. They find that underlying LLMs, when provided with textual scene descriptions, consistently outperform VLMs that use visual embeddings. Their analysis shows that this performance gap is due to VLMs' difficulty in translating visual information into accurate representations for reasoning. Our work builds on this insight by developing a modular approach to bridge this gap without requiring extensive retraining of VLMs.

Concurrently, \citet{chow2025physbench} introduce PhysBench, a comprehensive benchmark for evaluating physical understanding in VLMs. Their work reveals that VLMs' physical reasoning does not scale proportionally with model size, training data, or input frame count. They identify perceptual and knowledge gaps as the primary sources of errors and propose using vision foundation models like Depth Anything~\cite{yang2024depth} and SAM~\cite{kirillov2023segment} to enhance visual perception. While our approach shares the goal of improving physical reasoning, we take a different direction by leveraging simulation data to train specialized context builders rather than relying on generic vision foundation models.

\subsection{Simulation for Physical Understanding}
Simulation has long been recognized as a valuable tool for teaching machines about physical dynamics \cite{battaglia2013simulation}. Previous approaches have incorporated simulation directly into the inference pipeline \cite{wu2015galileo,wu2017learning,liu2022mind}, requiring computationally expensive simulators at inference time. In contrast, our method leverages simulation only during the training phase to generate rich physical descriptions, eliminating the need for simulation during inference while still benefiting from the detailed annotations that simulations provide. 

\subsection{Enhancing General Capabilities of VLMs}
Recent work has explored modular approaches to enhance VLM capabilities without full model retraining. Chain-of-thought prompting techniques \cite{wei2022chain} leverage large language models' reasoning abilities by encouraging step-by-step thinking, while multimodal chain-of-thought approaches \cite{zhang2023multimodal} extend this to vision-language tasks. Our Physics Context Builders build on these ideas by creating modular components that specialize in translating visual inputs into detailed physical descriptions.

\section{Understanding the Effect of Training Data on Physical Reasoning}
\label{sec:training-data}
\begin{figure}[t]
     \centering
    \includegraphics[width=0.9\linewidth]{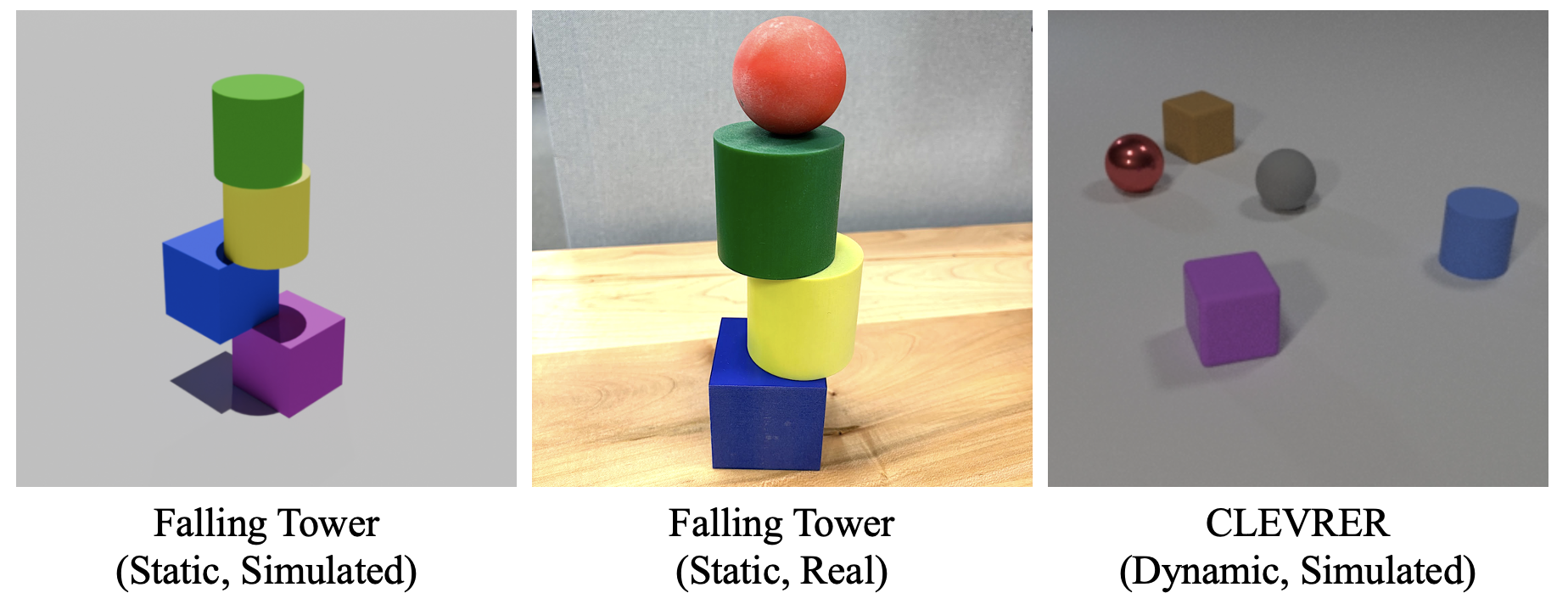}
    \caption{Datasets used for experiments: Falling Tower representing static physics in simulated and real environments; CLEVRER representing dynamic physics in a simulated environment.}
    \label{fig:dataset}   
\end{figure}
We begin by investigating why current VLMs struggle with physical reasoning. After outlining our experimental setup, covering datasets, models, and training procedures, we present results showing how fine-tuning with physics-focused data significantly improves performance. We then analyze the trade-off between data quantity and quality.

\subsection{Experimental Setup}

\paragraph{Datasets.} We utilize two benchmarks: CLEVRER~\cite{yi2019clevrer} for dynamic physical reasoning and Falling Tower for static stability detection (\cref{fig:dataset}).
\begin{itemize}[leftmargin=*]
    \item \textbf{CLEVRER}~\cite{yi2019clevrer} is an established benchmark containing 10,000 training videos, 5,000 validation videos, and 5,000 test videos, each paired with multiple questions across four categories: descriptive, explanatory, predictive, and counterfactual. The training set consists of 109,952 descriptive, 16,799 explanatory, 7,179 predictive, and 18,642 counterfactual questions. All descriptive questions are open-ended. Other question types are multiple-choice with up to four options and may have multiple correct answers. We report the accuracy of the models stratified by the question types. For multi-choice questions, we report both per-option accuracy, which measures the model’s overall correctness on single options across all questions, and per-question accuracy that measures the model’s performance on full questions, requiring all the choices to be selected correctly. Since the test set answers are not publicly available, we evaluate on the validation set. 
    
    \item \textbf{Falling Tower} is a dataset we create to complement CLEVRER by focusing on static physics. Similar to ShapeStacks~\cite{groth2018shapestacks} and block towers in \cite{lerer2016learning}, it features stacked objects but extends previous work by including question-answer pairs and simulation-generated annotations. The dataset contains 4,864 images of stacked objects (with 15 object types across 3 shapes and 5 colors) and 72,775 QA pairs, with 75\% used for training and 25\% for evaluation. Questions and  corresponding answers are generated by applying transformation functions to the simulation annotations, converting it into natural language questions and calculating the answers. For instance, to assess the stability of a stacked tower of objects, we compare the initial and final positions of the objects to determine if significant movement occurred. If an object remains within a predefined threshold, it is considered stable. Questions are then categorized into descriptive (e.g., "How many objects are in the scene?") and stability (e.g., "Will this collection of objects remain stationary?"). This dataset also enables the evaluation of Sim2Real transfer; we include 20 real-world images captured with 3D-printed objects with 100 human-generated QAs. For more details on Falling Tower, see \cref{appx:falling-tower}.
\end{itemize}

\paragraph{Models and evaluation.} We evaluate several zero-shot baselines, including GPT-4o, GPT-4o-mini~\cite{achiam2023gpt,hurst2024gpt}, Gemini-1.5-Pro~\cite{team2024gemini}, and PaliGemma-3B-mix (a variant fine-tuned on academic datasets)~\cite{beyer2024paligemma} with chain-of-thought prompting to encourage explicit reasoning. To study training data effects, we fine-tune PaliGemma-3B on CLEVRER and Falling Tower independently using Low-Rank Adaptation (LoRA)~\cite{hu2021lora}, minimizing the auto-regressive negative log-likelihood of the answers, conditioned on the questions and input video/image. For videos, we sample 8 frames and append them to the input context, except for the Gemini model, where we use the entire video as the input since they support native video processing. For all models, we take the final answer after the reasoning chain for evaluation. \cref{appx:qa-training} provides more details on the training procedure.

\paragraph{Framing the questions.} For all questions, including the open-ended and multi-choice ones, we provide the potential options in the question statement. For CLEVRER, we reframe multi-choice questions as a set of binary questions, asking whether each option is a valid answer. This yields significant improvement in the accuracy of all the evaluated models, as reported in \cref{appx:multi-choice}.

\subsection{Is Physics-Focused Training Data Enough?}

\begin{table*}[t]
\centering
\caption{Performance of zero-shot VLMs and the fine-tuned model on the Falling Tower benchmark. The second number after the first slash is the Sim2Real accuracy.}
\begin{adjustbox}{width=0.9\linewidth,center}
{
\begin{tabular}{llccccc}
\toprule
\textbf{Category} & \textbf{Model} & \multicolumn{3}{c}{\textbf{Descriptive} [sim acc. / real acc.]} & \multicolumn{2}{c}{\textbf{Stability} [sim acc. / real acc. / real F1]} \\
\cmidrule(lr){3-5} \cmidrule(lr){6-7}
& & num. obj. & num. obj. $\uparrow\downarrow$ & obj. $\uparrow\downarrow$ & obj. stable & tower stable \\
\midrule
\textbf{Random} & Random & 33.3 & 25.0 & 14.3 & 50.0 & 50.0 \\
\midrule
\multirow{5}{*} {\textbf{\shortstack{Zero-shot CoT}}} 
& GPT-4o & 99.3 / \textbf{100.0} & 91.4 / \textbf{100.0} & 99.4 / \textbf{95.0} & 56.9 / 60.0 / 63.3 & 59.6 / 55.0 / 54.2\\
& GPT-4o-mini & 94.9 / 89.5 & 61.3 / 84.2 & 87.9 / 73.7 & 49.0 / 52.6 / 55.0 & 53.1 / 36.8 / 19.8\\
& Gemini 1.5 Pro  & 97.2 / 95.0 & 89.9 / \textbf{100.0} & 97.8 / \textbf{95.0} & 54.6 / \textbf{80.0} / \textbf{80.0} & 60.5 / 60.0 / 60.1\\
& PaliGemma-3B-mix & 91.4 / 90.0 & 44.5 / 70.0 & 73.0 / 65.0 & 51.0 / 75.0 / 68.6 & 39.1 / \textbf{65.0} / 51.2\\
\midrule
\textbf{Fine-tuned QA}
& PaliGemma-3B-Fine-Tuned & \textbf{100.0 / 100.0} & \textbf{100.0 / 100.0} & \textbf{100.0 / 95.0} & \textbf{84.6} / 70.0 / 73.0 & \textbf{87.6} / \textbf{65.0} / \textbf{64.4}\\
\bottomrule
\end{tabular}
}
\end{adjustbox}
\label{tab:tower}
\end{table*}

\begin{table*}[t]
\centering
\caption{Performance of zero-shot VLMs and the fine-tuned model on the CLEVRER benchmark.}
\begin{adjustbox}{width=0.9\linewidth,center}
{
\begin{tabular}{llccccccc}
\toprule
\textbf{Category} & \textbf{Model} & {\textbf{Descriptive}} & \multicolumn{2}{c}{\textbf{Explanatory}} & \multicolumn{2}{c}{\textbf{Predictive}}& \multicolumn{2}{c}{\textbf{Counterfactual}}\\
\cmidrule(lr){4-5} \cmidrule(lr){6-7} \cmidrule(lr){8-9}
& & & per ques. & per opt. & per ques. & per opt. & per ques. & per opt.\\
\midrule
\textbf{Random} & Random &28.8 & 11.8 & 50.0 & 25.0 & 50.0 & 7.4 & 50.0 \\
\midrule
\multirow{5}{*}{\textbf{\shortstack{Zero-shot CoT }}} 
& GPT-4o & 62.7 & 30.7 & 65.5& 30.3 & 47.6 & 18.7 & 60.2\\
& GPT-4o-mini & 49.5 &9.3 & 51.8 &44.9 & 45.1&15.6 & 51.0\\
& Gemini 1.5 Pro & 58.6 & 15.7 & 61.2 & 32.0 & 49.6 & 17.6 & 55.6 \\
& PaliGemma-3B-mix & 38.9 & 6.6 & 33.4 & 44.7 & 48.7 & 7.7 &49.8 \\
\midrule
\textbf{Fine-tuned QA} & {PaliGemma-3B-Fine-Tuned} & \textbf{92.9}& \textbf{94.7}  & \textbf{98.2} & \textbf{83.6} & \textbf{83.6} & \textbf{68.4} & \textbf{88.7}\\
\bottomrule
\end{tabular}
}
\end{adjustbox}
\label{tab:clevrer}
\end{table*}

\cref{tab:tower,tab:clevrer} present our results for the Falling Tower and CLEVRER benchmarks. Several key insights emerge:

\paragraph{Zero-shot VLMs struggle with physical reasoning despite strong descriptive capabilities.} On Falling Tower, large models like GPT-4o and Gemini-1.5-Pro demonstrate near-perfect accuracy (95-100\%) on descriptive questions, both in simulated and real environments. However, their performance drops substantially on stability questions, with accuracy ranging from only 55-60\% (barely above random guessing). Similarly, on CLEVRER, while GPT-4o achieves 62.7\% accuracy on descriptive questions, its performance falls to 30.7\%, 30.3\%, and 18.7\% on explanatory, predictive, and counterfactual questions, respectively. This substantial performance gap highlights that while VLMs have developed strong capabilities for \emph{high-level} scene understanding and description, they lack the abilities needed to predict outcomes or explain causal relationships accurately.

\paragraph{Fine-tuning with physics-focused data substantially improves physical reasoning.} Unsurprisingly, the fine-tuned PaliGemma-3B model shows dramatic improvements across all tasks. On Falling Tower, it achieves perfect descriptive accuracy (100\%) and substantial gains in stability prediction (up to 28\% compared to GPT-4o). On CLEVRER, the improvements are even more significant, reaching close to the state-of-the-art results by Aloe, a specialized architecture trained on CLEVRER (Descriptive: 94.0 \%, Explanatory: 96.0\%, Predictive: 87.5\%, Counterfactual: 75.6\%)~\cite{ding2021attention}. These results demonstrate that targeted fine-tuning with physics-focused data can significantly enhance physical reasoning capabilities, enabling a 3B parameter model to outperform much larger state-of-the-art models. See \cref{appx:spec-clevrer} for a comparison to other specialized baselines.

\paragraph{Sim2Real transfer is successful.} The fine-tuned model maintains strong performance when transferring from simulated to real-world images in Falling Tower. It achieves 100\% accuracy on descriptive tasks and 70.0\% / 65.0\% accuracy on object / tower stability questions, respectively. 

\subsection{Ablation Studies: Not All Data is Equal}

To gain deeper insights into how training data composition affects physical reasoning capabilities, we conduct several ablation studies. We provide more ablations in \cref{appx:ablations}.

\begin{table}[t]
\centering
\caption{Effect of fine-tuning PaliGemma-3B on different QA types in Falling Tower.}
\begin{adjustbox}{width=\linewidth,center}
{\Huge
\begin{tabular}{lccccc}
\toprule
\textbf{Fine-tuning Data} & \multicolumn{3}{c}{\textbf{Descriptive} [acc.]} & \multicolumn{2}{c}{\textbf{Stability} [acc.]} \\
\cmidrule(lr){2-4} \cmidrule(lr){5-6}
 & num. obj. & num. obj. $\uparrow\downarrow$ & obj. $\uparrow\downarrow$ & obj. stable & tower stable \\
\midrule
{\textbf{No Fine-tuning}} & 50.9 & 21.2 & 8.5 & 51.0 & 39.1 \\
\midrule
{\textbf{Stability QAs}} & {52.4} & {26.0} & {41.2} & {84.4} & {86.5} \\
\midrule
{\textbf{Descriptive QAs}} & {100.0} & {100.0} & {100.0} & {51.0} & {39.1} \\
\midrule
{\textbf{All QAs}} & {100.0} & {100.0} & {100.0} & {84.6} & {87.6} \\
\bottomrule
\end{tabular}
}
\end{adjustbox}
\label{tab:finetune-qa}
\end{table}

\paragraph{Specificity of QA types in fine-tuning.} \cref{tab:finetune-qa} shows the impact of using different subsets of QA pairs for fine-tuning on Falling Tower. Fine-tuning with only descriptive questions achieves perfect descriptive accuracy but fails to improve stability prediction (remaining at the baseline level). Interestingly, fine-tuning with only stability questions not only improves stability prediction substantially but also yields moderate improvements in descriptive tasks, particularly for object identification (41.2\% vs. 8.5\% baseline). However, the best performance comes from fine-tuning on the full QA dataset. This highlights the need to carefully select data to cover all QA types, which can be seen as a limitation of targeted fine-tuning in VLMs.

\begin{figure}[t]
\centering
\includegraphics[width=\linewidth]{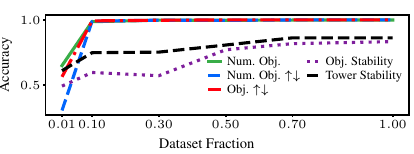}
\caption{The amount of training data vs. performance on different QA types in Falling Tower.}
\label{fig:finetune-qa}
\end{figure}

\paragraph{Data efficiency varies by task type.} \cref{fig:finetune-qa} illustrates how accuracy varies with the amount of training data for different QA types in Falling Tower. Descriptive questions reach near-saturation with only $\sim$10\% of the data, showing the model can quickly learn basic scene understanding. In contrast, stability questions benefit more from extra data, though with diminishing returns. Even with 100\% of the available data, the model does not reach perfect performance on stability questions (87.6\% for tower stability). This shows that increasing dataset size alone is insufficient; improvements in physical reasoning also require diverse, targeted supervision.

\section{Physics Context Builders: A Modular Approach to Physical Reasoning}

While fine-tuning with physics-focused data can significantly enhance physical reasoning in VLMs, it presents notable practical limitations. Fine-tuning state-of-the-art foundation models like GPT-4o or Gemini is often expensive or even impossible due to their closed-source nature and computational requirements. Furthermore, the task-specific nature of fine-tuning, as seen in \cref{sec:training-data}, means separate models may be needed for different physical reasoning tasks, which limits the generalizability of the approach.
To address these challenges, we introduce \emph{Physics Context Builders} (PCBs) -- a modular and efficient approach that enhances physical reasoning capabilities of existing VLMs without modifying them directly. PCBs leverage the strong in-context learning abilities of large language models~\cite{brown2020language,wei2022chain}, which have shown impressive performance with textual scene descriptions~\cite{nagar2024zero}.

\subsection{Methodology}

PCBs are specialized VLMs fine-tuned on simulation data to generate detailed physical descriptions of visual scenes. Rather than directly answering questions, PCBs act as perception modules that translate visual inputs into rich textual descriptions capturing the physical properties and dynamics of a scene. These descriptions then serve as enhanced context for larger VLMs, enabling more accurate physical reasoning through in-context learning. \cref{fig:data_gen_diagram} provides an overview of our approach. PCBs offer several advantages:
\begin{itemize}[leftmargin=*]
    \item \textbf{Modularity:} PCBs can be fine-tuned and deployed independently of the foundation model.
    \item \textbf{Efficiency:} Only smaller, specialized models need to be fine-tuned, rather than larger VLMs. 
    \item \textbf{Flexibility:} Different PCBs can be developed for different physical phenomena, creating a toolbox of physical reasoning enhancers.
    \item \textbf{Compatibility:} PCBs work with any (vision) language model capable of in-context learning, including closed-source commercial models.
\end{itemize}

\subsubsection{Physical Context Generation}

Using annotations from the simulator, PCBs generate question-agnostic descriptions that capture the physical essence of a scene. We consider two context types:

\noindent \textbf{(1) Human-like Narration (HN):} Produces intuitive natural language descriptions of the scene's physical properties and events that can be more aligned with foundation models:
    
    \begin{figure}[ht]
        \centering
        \includegraphics[width=\linewidth]{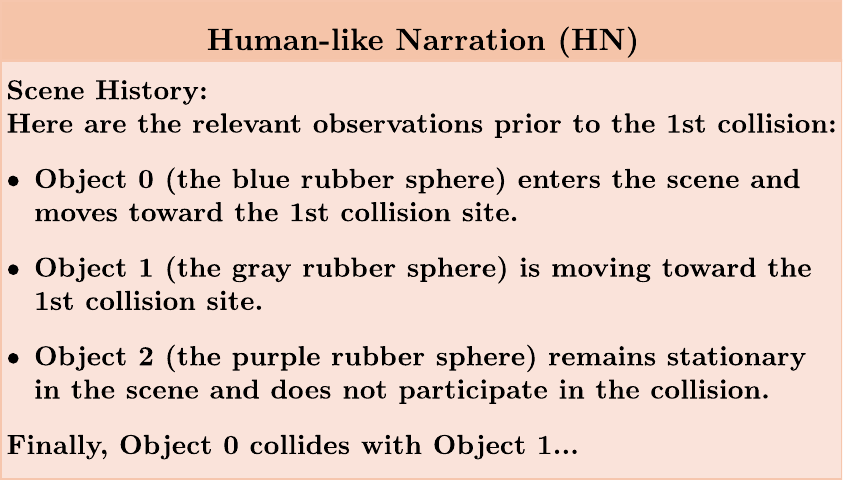}
    \end{figure}
    
\noindent \textbf{(2) Structured Physics (SP):} Provides frame-by-frame structured observations in a format similar to a physics simulation output, with standardized tags for physical properties. This structured approach helps models capture precise temporal relationships and physical properties:

    \begin{figure}[ht]
        \centering
        \includegraphics[width=\linewidth]{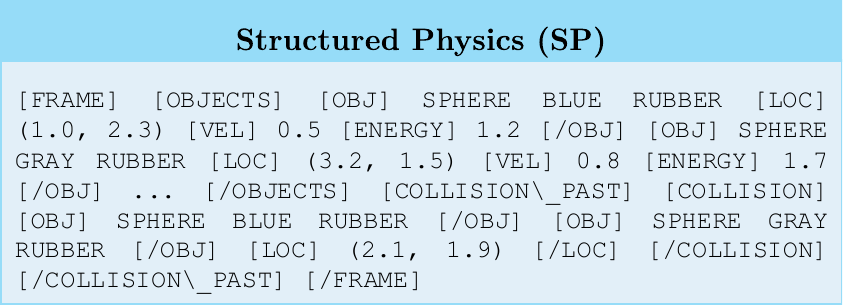}
    \end{figure}

\subsubsection{Training PCBs}

We train PCBs by fine-tuning a pre-trained VLM to generate physical descriptions from visual inputs. Specifically:

\begin{enumerate}[leftmargin=*]
    \item We use PaliGemma-3B as the base pre-trained VLM due to its strong vision-language capabilities and reasonable computational requirements. 
    \item We apply LoRA-based fine-tuning~\cite{hu2021lora}, minimizing the auto-regressive negative log-likelihood of the context descriptions conditioned on the input video/image. See \cref{appx:pcb-training} for more details.
    \item For videos (e.g., in CLEVRER), we sample 8 frames and append them to the input context.
    \item Fine-tuning jointly optimizes both vision and language components to ensure alignment between visual features and linguistic representations.
\end{enumerate}

The training data is generated from the same simulations used to create the QA pairs, but focuses on comprehensive scene descriptions rather than specific questions and answers. This approach enables PCBs to provide rich context independent of the specific reasoning task.

\subsection{Experimental Results}

We evaluate the effectiveness of PCBs by integrating them with three foundation models: GPT-4o, GPT-4o-mini, and Gemini-1.5-Pro. In each experiment, we first pass the visual input to the appropriate PCB, which generates a physical description. This description is then provided as additional context to the foundation model along with the user's question and the video/image.

\begin{table*}[t]
\centering
\caption{Performance of foundation models augmented with Physics Context Builders (PCBs) on the Falling Tower benchmark. HN refers to Human Narration style PCB. The numbers in parentheses indicate improvements over the respective zero-shot baselines. The second number after the slash is the Sim2Real accuracy and the third number after the second slash is the F1 score on Sim2Real.}
\begin{adjustbox}{width=0.9\linewidth,center}
{\Huge
\begin{tabular}{llccccc}
\toprule
\textbf{Category} & \textbf{Model} & \multicolumn{3}{c}{\textbf{Descriptive} [sim acc. / real acc.]} & \multicolumn{2}{c}{\textbf{Stability} [sim acc. / real acc. / real F1]} \\
\cmidrule(lr){3-5} \cmidrule(lr){6-7}
& & num. obj. & num. obj. $\uparrow\downarrow$ & obj. $\uparrow\downarrow$ & obj. stable & tower stable \\
\midrule
\multirow{3}{*}{\textbf{Zero-shot CoT}} 
& GPT-4o & 99.3 / \textbf{100.0} & 91.4 / \textbf{100.0} & 99.4 / \textbf{95.0} & 56.9 / 60.0 / 63.3 & 59.6 / 55.0 / 54.2\\
& GPT-4o-mini & 94.9 / 89.5 & 61.3 / 84.2 & 87.9 / 73.7 & 49.0 / 52.6 / 55.0 & 53.1 / 36.8 / 19.8\\
& Gemini 1.5 Pro & 97.2 / 95.0 & 89.9 / \textbf{100.0} & 97.8 / \textbf{95.0} & 54.6 / \textbf{80.0} / \textbf{80.0} & 60.5 / 60.0 / 60.1\\
\midrule
\multirow{6}{*}{\textbf{VLM + PCB (HN)}} 
& GPT-4o-PCB & 99.5 / \textbf{100.0} & {\bf 97.6} / \textbf{100.0} & {\bf 99.5} / \textbf{95.0} & {\bf 76.7} / {75.0 / 73.8} & {\bf 85.1} / \textbf{70.0 / 65.6} \\
& & (+0.2) / (0.0) & (+6.2) / (0.0) & (+0.1) / (0.0) & (+19.8) / (+15.0) / (+10.5) & (+25.5) / (+15.0) / (+11.4) \\
& GPT-4o-mini-PCB & {\bf 99.9} / 95.0 & 74.3 / 90.0 &97.5 / \textbf{95.0} & 75.0 / 70.0 / 73.0 & 84.7 / 40.0 / 33.8 \\
& & (+5.0) / (+5.5) & (+13.0) / (+5.8) & (+9.6) / (+21.3) & (+26.0) / (+17.4) / (+18.0) & (+31.6) / (+3.2) / (+14.0)\\
& Gemini 1.5 Pro-PCB & 97.9 / \textbf{100.0} & 97.5 / \textbf{100.0} & 97.4 / 94.7 & 75.9 / 73.7 / 76.7 & 84.9 / 57.9 / 59.1 \\
& & (+0.7) / (+5.0) & (+7.6) / (0.0) & (-0.4) / (-0.3) & (+21.3) / (-6.3) / (-3.3) & (+24.4) / (-2.1) / (-1.0)\\
\bottomrule
\end{tabular}
}
\end{adjustbox}
\label{tab:tower-pcb}
\end{table*}

\begin{table*}[t]
\centering
\caption{Performance of foundation models augmented with Physics Context Builders (PCBs) on the CLEVRER benchmark. HN is Human Narration and SP is Structured Physics. The numbers in parentheses indicate improvements over the respective zero-shot baselines.}
\begin{adjustbox}{width=0.65\linewidth,center}
{\Huge
\begin{tabular}{llccccc}
\toprule
\textbf{Category} & \textbf{Model} & {\textbf{Descriptive}} & \multicolumn{2}{c}{\textbf{Explanatory}} & \multicolumn{2}{c}{\textbf{Counterfactual}}\\
\cmidrule(lr){4-5} \cmidrule(lr){6-7} 
& & & per ques. & per opt. & per ques. & per opt.\\
\midrule
\multirow{3}{*}{\textbf{\shortstack{Zero-shot CoT }}} 
& GPT-4o & 62.7 & 30.7 & 65.5 & 18.7 & 60.2\\
& GPT-4o-mini & 49.5 &9.3 & 51.8 &15.6 & 51.0\\
& Gemini 1.5 Pro & 58.6 & 15.7 & 61.2 & 17.6 & 55.6 \\
\midrule
\multirow{3}{*} {\textbf{\shortstack{VLM + PCB\\(HN)}}} 
& GPT-4o-PCB & {\bf 75.6} (+12.9) & {\bf 41.6} (+10.9) & 67.0 (+1.5) & {\bf 28.2} (+9.5) & {\bf 68.4} (+8.2) \\
& GPT-4o-mini-PCB & 65.7 (+16.2) & 26.8 (+17.5) & 62.2 (+10.4)  & 17.3 (+1.7) & 52.8 (+1.8) \\
& Gemini 1.5 Pro-PCB & 72.8 (+14.2) & 35.6 (+19.9) & {\bf 70.8} (+9.6) & 26.2 (+8.6) & 64.9 (+9.3)\\
\midrule
\multirow{3}{*} {\textbf{\shortstack{VLM + PCB\\(SP)}}} 
& GPT-4o-PCB & 70.0 (+7.3) & 34.9 (+4.2) & 61.1 (-4.4) & 19.2 (+0.5) & 63.3 (+3.1) \\
& GPT-4o-mini-PCB & 58.6 (+9.1) & 19.2 (+9.9) & 58.9 (+7.1) &16.2 (+0.6)& 51.8 (+0.8)\\
& Gemini 1.5 Pro-PCB & 67.4 (+8.8)& 30.5 (+14.8)& 68.7 (+7.5) & 21.1 (+3.5)& 60.5 (+4.9) \\
\bottomrule
\end{tabular}
}
\end{adjustbox}
\label{tab:clevrer-pcb}
\end{table*}

\subsubsection{PCB Performance on Falling Tower}

\cref{tab:tower-pcb} shows the performance of foundation models augmented with PCBs on Falling Tower. For additional experiments, see \cref{appx:add-intern}. Several key findings:

\begin{itemize}[leftmargin=*]
    \item \textbf{Substantial improvement in stability tasks:} PCBs provide remarkable gains in stability prediction, with GPT-4o-PCB showing up to 25.5\% improvement in tower stability assessment. These gains are even more pronounced for smaller models like GPT-4o-mini, which sees a 31.6\% improvement in tower stability prediction.
    
    \item \textbf{Modest gains in descriptive tasks:} Since foundation models already perform well in descriptive tasks, PCBs offer limited additional benefits in this area, with improvements mainly in the more challenging descriptive tasks like identifying objects above/below others (e.g., 13.0\% improvement for GPT-4o-mini).
    
    \item \textbf{Effective Sim2Real transfer:} PCB-augmented models show improved generalization to real-world scenarios, with GPT-4o-PCB achieving a 15.0\% gain in real-world tower stability prediction compared to zero-shot.
    
    \item \textbf{Model size effects:} Interestingly, the smaller GPT-4o-mini model shows greater relative improvements when augmented with PCBs compared to larger ones, possibly due to its more limited perception.
\end{itemize}

\subsubsection{PCB Performance on CLEVRER}

\cref{tab:clevrer-pcb} presents results for CLEVRER, a more challenging benchmark that requires dynamic physical reasoning:

\begin{itemize}[leftmargin=*]
    \item \textbf{Human Narration vs. Structured Physics:} Human-like narration (HN) consistently outperforms structured physics (SP) descriptions across all models and question types. This can be due to foundation models' better understanding of natural language descriptions compared to more structured, technical formats.
    
    \item \textbf{Strong improvements in descriptive and explanatory tasks:} PCBs provide substantial gains in descriptive accuracy (up to 16.2\% for GPT-4o-mini) and explanatory reasoning (up to 19.9\% for Gemini 1.5 Pro).
    
    \item \textbf{Limited gains in counterfactual reasoning:} While PCBs improve counterfactual reasoning (1.7–9.5\% gains), the improvements are more modest, reflecting the intrinsic complexity of this task, even with enriched context.
\end{itemize}

\noindent\textbf{Remark.} For CLEVRER, we omit PCB evaluation on predictive questions since our current PCBs are designed to describe observed scenes rather than predict future events. Future work could address this by developing predictive PCBs that generate plausible future scene descriptions based on simulation rollouts.

\subsection{Multi-Agent Framework for PCB Integration}
\begin{figure}[t]
\centering
\includegraphics[width=\linewidth]{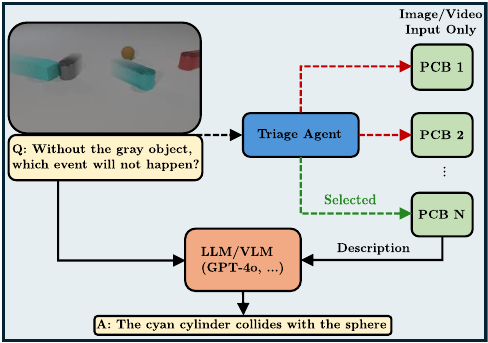}
\caption{Multi-agent framework for PCB integration. A triage agent selects the appropriate PCB based on the user query, and the PCB generates a detailed scene description that enriches the foundation model’s context.}
\label{fig:cb-diagram}
\end{figure}
Given the ability of foundational models to interpret the overall context of scenes effectively, we explore whether they can reliably select the appropriate PCB when provided with a question-image pair. We utilize a multi-agent triage model inspired by OpenAI's Swarm architecture~\cite{githubGitHubOpenaiswarm}. As illustrated in \cref{fig:cb-diagram}, our multi-agent framework consists of:

\begin{enumerate}[leftmargin=*]
    \item A \textbf{Triage Agent} that analyzes the user query and visual input to identify the required type of physical reasoning.
    \item Multiple \textbf{PCBs}, each specialized for different physical phenomena (e.g., stability analysis, collision detection, motion tracking).
    \item A \textbf{Foundation Model} that receives the PCB-generated context and the original query to produce the response.
\end{enumerate}

In our evaluation, each input consists of a natural language question from a QA dataset paired with a corresponding scene image. This input is initially processed by GPT-4o or GPT-4o-mini that routes the query to one of two specialized PCBs: the PCB for the Falling Tower dataset, or PCB designed for analyzing motion and object interactions as in the CLEVRER dataset. As shown in \cref{tab:performance_comparison_agent}, both GPT-4o and GPT-4o-mini achieve excellent F1-Scores, effectively routing queries to the appropriate PCB. These results suggest that PCBs in a multi-agent framework offer a promising approach, with foundation models capable of reliably selecting the correct PCB based on the question-image pair.

\begin{table}[t]
\caption{Accuracy and F1-scores for selecting the correct Physics Context Builder using the triage agent in a two-agent setup.}
\begin{adjustbox}{width=0.9\linewidth,center}
{
\begin{tabular}{lcc}
\toprule
\textbf{Task} & \textbf{Metric} & \textbf{GPT-4o / GPT-4o-mini} \\
\midrule
\multirow{2}{*}{Stacked Objects} & Accuracy (\%) & 87.67 / 97.00 \\
& F1-Score & 0.9326 / 0.9782 \\
\midrule
\multirow{2}{*}{Dynamic Scene} & Accuracy (\%) & 94.67 / 98.67 \\
& F1-Score & 0.9403 / 0.9785 \\
\bottomrule
\end{tabular}
}
\end{adjustbox}
\label{tab:performance_comparison_agent}
\end{table}

\section{Discussion}

Our results demonstrate that Physics Context Builders (PCBs) offer a promising approach to enhancing physical reasoning in Vision-Language Models (VLMs). PCBs increase the average performance of GPT-4o (GPT-4o-mini) by 11.1\% (11.8\%) on CLEVRER and 8.2\% (13.8\%) on Falling Tower. These improvements are particularly notable given that no modification to the foundation models was required. Despite these gains, large VLMs still exhibit sub-optimal performance in tasks requiring deeper reasoning, such as counterfactual questions (with improvements limited to 1.7-9.5\%) and explanatory questions. While enhancing visual perception through PCBs is important, especially for descriptive tasks, the subpar performance on more involved reasoning tasks calls for additional interventions to achieve a comprehensive physical understanding.

Besides the performance gain, our findings confirm the value of simulation data in addressing the limitations of VLMs. Unlike simulation-in-the-loop approaches \cite{wu2015galileo,wu2017learning,liu2022mind,ding2021dynamic}, which require computationally expensive simulators during inference, our method uses simulation only to generate synthetic data (context descriptions) for fine-tuning smaller VLMs. This approach is more efficient at inference while still leveraging the rich annotations that simulations can provide. The effectiveness of PCBs in generalizing from simulated to real-world data further supports this approach, as demonstrated by the successful Sim2Real transfer in the Falling Tower benchmark.

PCBs provide a modular and efficient framework for enhancing the physical reasoning capabilities of foundation models without requiring direct fine-tuning. Generating rich physical context from visual inputs enhances the perceptual ability of foundation models, resulting in more accurate reasoning across diverse physical tasks, from static stability to dynamic collision detection. The multi-agent framework further enhances this approach by enabling adaptive selection of specialized PCBs based on the specific reasoning task.

\section{Limitations and Future Work}

While our work demonstrates the effectiveness of PCBs across benchmarks, several limitations remain. First, our current benchmarks focus on a relatively constrained set of physical phenomena—primarily rigid body dynamics and stability. This limits our ability to evaluate how well these approaches generalize to the full spectrum of physical reasoning that humans perform intuitively, such as fluid dynamics or object manipulation.

Second, our framework requires annotated data, which naturally comes from simulation. For unannotated videos, such as those available on YouTube, the lack of structured annotations presents a challenge. An important open problem is how PCBs could extract detailed physical descriptions from real-world videos without explicit annotations, potentially through self-supervised learning or by leveraging other foundation models to generate pseudo-annotations.

Third, the performance improvements, while substantial, still leave room for further enhancement, particularly for complex reasoning tasks like counterfactual/predictive questions. Since PCBs inherently perform direct translation from visual signals to text, they lead to strong descriptive capabilities in the foundation models. However, this translation process does not capture all the visual cues needed for more complex reasoning. Hence, we hypothesize that more comprehensive textual descriptions, possibly containing counterfactual/predictive information, could significantly improve the performance of PCBs on these challenging tasks.

A promising future direction is to tackle more complex physical reasoning tasks that require larger-scale simulations. This might involve integrating advanced physics engines that simulate phenomena like fluid dynamics, deformable materials, and articulated mechanisms. Notably, while these simulations could become computationally expensive for simulation-in-the-loop approaches during inference, our PCB framework uses simulation data only for fine-tuning and therefore does not add significant computational burden at inference time.

Another avenue for future work is to explore how multiple PCBs could be composed or chained together to handle scenarios requiring reasoning about multiple physical phenomena simultaneously. For instance, reasoning about a scene involving both rigid body dynamics and fluid interactions might benefit from specialized PCBs for each phenomenon, with their outputs combined to provide comprehensive context to the foundation model.

\section*{Acknowledgements}
{
We thank Brian Jeong for 3D printing the parts used in the Falling Tower dataset. Resources for this research were primarily provided by Autodesk Research, with partial support from the Province of Ontario, the Government of Canada through CIFAR, and sponsors of the Vector Institute.
}

{
    \bibliographystyle{ieeenat_fullname}
    \bibliography{main}
}

\appendix
\input supplementary

\end{document}

%% file: math_commands.tex
\newtheorem*{rep@theorem}{\rep@title}
\newcommand{\newreptheorem}[2]{%
\newenvironment{rep#1}[1]{%
 \def\rep@title{#2 \ref{##1}}%
 \begin{rep@theorem}}%
 {\end{rep@theorem}}}
\makeatother

\newreptheorem{theorem}{Theorem}

\newreptheorem{corollary}{Corollary}

\newreptheorem{proposition}{Proposition}

\theoremstyle{definition}

\def\ddefloop#1{\ifx\ddefloop#1\else\ddef{#1}\expandafter\ddefloop\fi}
\def\ddef#1{\expandafter\def\csname bb#1\endcsname{\ensuremath{\mathbb{#1}}}}
\ddefloop ABCDEFGHIJKLMNOPQRSTUVWXYZ\ddefloop
\def\ddefloop#1{\ifx\ddefloop#1\else\ddef{#1}\expandafter\ddefloop\fi}
\def\ddef#1{\expandafter\def\csname b#1\endcsname{\ensuremath{\mathbf{#1}}}}
\ddefloop ABCDEFGHIJKLMNOPQRSTUVWXYZ\ddefloop
\def\ddef#1{\expandafter\def\csname c#1\endcsname{\ensuremath{\mathcal{#1}}}}
\ddefloop ABCDEFGHIJKLMNOPQRSTUVWXYZ\ddefloop
\def\ddef#1{\expandafter\def\csname h#1\endcsname{\ensuremath{\hat{#1}}}}
\ddefloop ABCDEFGHIJKLMNOPQRSTUVWXYZ\ddefloop
\def\ddef#1{\expandafter\def\csname wh#1\endcsname{\ensuremath{\widehat{#1}}}}
\ddefloop abcdefghijklmnopqrstuvwxyz\ddefloop
\def\ddef#1{\expandafter\def\csname hc#1\endcsname{\ensuremath{\widehat{\mathcal{#1}}}}}
\ddefloop ABCDEFGHIJKLMNOPQRSTUVWXYZ\ddefloop
\def\ddef#1{\expandafter\def\csname t#1\endcsname{\ensuremath{\widetilde{#1}}}}
\ddefloop ABCDEFGHIJKLMNOPQRSTUVWXYZ\ddefloop
\def\ddef#1{\expandafter\def\csname tc#1\endcsname{\ensuremath{\widetilde{\mathcal{#1}}}}}
\ddefloop ABCDEFGHIJKLMNOPQRSTUVWXYZ\ddefloop

\usepackage{prettyref}
\usepackage{xcolor}

\newcommand{\xhdr}[1]{\textbf{#1.}}

%% file: supplementary.tex
\onecolumn
\section{Training Details}
Here, we provide the training details, including the hyperparameters, for both QA fine-tuning and PCB training tasks. All training was performed on NVIDIA A100 DGX systems.
\subsection{QA fine-tuning}
\label{appx:qa-training}
To fine-tune PaliGemma-3B on the question-answer datasets, we apply the LoRA fine-tuning scheme by targeting the attention weights in both the vision and language modules, as well as the fully connected MLP layers, multi-modal projector layers, embedding tokens, patch embedding, and positional embedding. \cref{tab:qa-finetuning-detail} (left) shows other hyperparameters used for QA fine-tuning.

\begin{table}[h!]
    \centering
    \caption{Hyperparameters used to fine-tune the PaliGemma-3B models on the QA datasets (left) and as PCB modules (right).}
    {\small
        \begin{tabular}{lcc}
            \toprule
           {\textbf{Hyperparameter}} & \textbf{Falling Tower} & \textbf{CLEVRER} \\
            \midrule       
            LoRA rank & 16 & 16 \\
            Learning rate & 5e-5 & 5e-5 \\
            Batch size & 32 & 32\\
            Epochs & 10 & 3 \\
            Trainable parameters & 1.24 \% & 1.24 \% \\
            Number of frames & 1 & 8 \\
            Compute time & $\sim$ 3.5 hours & $\sim$ 37 hours \\
            \bottomrule
        \end{tabular}
        \quad 
        \begin{tabular}{lcc}
            \toprule
           {\textbf{Hyperparameter}} & \textbf{Falling Tower} & \textbf{CLEVRER} \\
            \midrule       
            LoRA rank & 16 & 16 \\
            Learning rate & 5e-5 & 5e-5 \\
            Batch size & 32 & 64\\
            Epochs & 50 & 10 \\
            Trainable parameters & 1.24 \% & 1.24 \% \\
            Number of frames & 1 & 8 \\
            Compute time & $\sim$ 1 hour & $\sim$ 2.3 hours \\
            \bottomrule
        \end{tabular}
        }
    \label{tab:qa-finetuning-detail}
\end{table}
\subsection{PCB training}
\label{appx:pcb-training}
\noindent \xhdr{Descriptions used for training PCBs} We first discuss the two types of descriptions we considered for training PCBs:
\begin{enumerate}[leftmargin=*]
\setlength\itemsep{1em}
    \item Human-Narration ({HN}), which generates a summary of all the collisions that occurred in the scene.\\
    
 \fbox{   
 \parbox{0.95\linewidth}{
 \small
 \texttt{Scene History:\\
In this scene, there are 3 collisions occurring in sequence.\\
Here are the relevant observations prior to the 1st collision:\\
Object 0  (the blue rubber sphere) enters the scene and moves toward the 1st collision site.\\
Object 1  (the gray rubber sphere) is moving toward the 1st collision site.\\
Object 2  (the cyan metal cube) enters the scene and is moving in the rest of the scene but does not participate in the collision.\\
Object 3  (the purple rubber sphere) remains stationary in the scene and does not participate in the collision.\\
Object 4  (the blue metal sphere) remains stationary in the scene and does not participate in the collision.\\
Finally, Object 0 collides with Object 1.\\
Here are the relevant observations prior to the 2nd collision:\\
...
}
}
}

    \item Structured-Physics ({SP}), which describes each provided video frame separately as follows, while adding physical properties of the objects, including their discretized and normalized locations and velocities. We also include the locations of collisions that occurred up to a certain frame. \\
 
             \fbox{   
 \parbox{0.95\linewidth}{
 \small
 \texttt{[FRAME] [OBJECTS]  [OBJ] SHAPE COLOR MATERIAL [LOC] LOC [/LOC] [VEL] VEL [/VEL] [/OBJ] [OBJ] SHAPE COLOR MATERIAL [LOC] LOC [/LOC] [VEL] VEL [/VEL] [/OBJ] ...  [/OBJECTS] [COLLISION\_PAST] [COLLISION] [OBJ] SHAPE COLOR MATERIAL [/OBJ]  [OBJ] SHAPE COLOR MATERIAL  [/OBJ] [LOC] LOC [/LOC] [/COLLISION] ... [/COLLISION\_PAST] [/FRAME]
}
}
}\\
\end{enumerate}
\noindent \xhdr{Training details} We use the pre-trained PaliGemma-3B model for training the PCB modules and apply the LoRA fine-tuning scheme, similar to the approach used for QA fine-tuning. \cref{tab:qa-finetuning-detail} (right) provides the hyperparameters used to train PCB modules for both Falling Tower and CLEVRER datasets.

\section{Ablations}
\label{appx:ablations}
\subsection{The Effect of Framing Multi-Choice Questions as Multiple Binary Questions}
\label{appx:multi-choice}
As discussed in the main paper, framing the multi-choice questions as multiple binary questions in CLEVRER can yield significant improvement in the accuracy of the models. In \cref{tab:multi-choice}, we provide a comparison between the performance of fine-tuned PaliGemma-3B models with and without this change. As demonstrated, we observe improvement in almost all categories, except for the per question predictive accuracy. We posit that this is because the predictive questions in CLEVRER are always binary questions with exactly one correct choice. Framing the predictive questions as two independent binary questions can result in a model choosing both options as correct or wrong. 

\begin{table}[h!]
    \centering
    \caption{The performance of the fine-tuned PaliGemma-3B model on question answer pairs for the CLEVRER benchmark based on framing the multi-choice questions as binary questions. Both models are trained for three epochs.}
        {\small
        \begin{tabular}{cccccccc}
            \toprule
            \textbf{Multi-Choice as Binary?} & {\textbf{Descriptive}} & \multicolumn{2}{c}{\textbf{Explanatory}} & \multicolumn{2}{c}{\textbf{Predictive}}& \multicolumn{2}{c}{\textbf{Counterfactual}}\\
            \cmidrule(lr){3-4} \cmidrule(lr){5-6} \cmidrule(lr){7-8} 
             & & per ques. & per opt. & per ques. & per opt. & per ques. & per opt.\\
            \midrule
             False & {89.3} & {69.0} & {86.6}  & \textbf{83.6}  & {83.6} & {41.0} & {74.0} \\ 
            \midrule
             True & \textbf{92.9} & \textbf{94.7} & \textbf{98.2} & 77.9 & \textbf{88.2} & \textbf{68.4}  & \textbf{88.7} \\ 
            \bottomrule
        \end{tabular}
        }
    \label{tab:multi-choice}
\end{table}

\subsection{The Effect of Training Epochs}
We run an ablation study to assess the effect of training for smaller vs. larger number of epochs on the accuracy of CLEVRER in the QA fine-tuning task. \cref{tab:epoch-clevrer} demonstrates a large improvement in training for more epochs.

\begin{table}[h!]
    \centering
    \caption{The performance of the fine-tuned PaliGemma-3B model on question answer pairs for the CLEVRER benchmark based on the number of trained epochs. Here, multi-choice questions are asked as they are (without framing them as multiple binary questions).}
        {\small
        \begin{tabular}{cccccccc}
            \toprule
            \textbf{Epochs} & {\textbf{Descriptive}} & \multicolumn{2}{c}{\textbf{Explanatory}} & \multicolumn{2}{c}{\textbf{Predictive}}& \multicolumn{2}{c}{\textbf{Counterfactual}}\\
            \cmidrule(lr){3-4} \cmidrule(lr){5-6} \cmidrule(lr){7-8} 
             & & per ques. & per opt. & per ques. & per opt. & per ques. & per opt.\\
            \midrule
             3 & \textbf{89.3} & \textbf{69.0} & \textbf{86.6}  & \textbf{83.6}  & \textbf{83.6} & \textbf{41.0} & \textbf{74.0} \\ 
            \midrule
             2 & {87.2} & {66.5} & {85.7}  & {82.3}  & {82.3} & {38.3} & {72.9} \\ 
            \midrule
             1 & {78.1} & {52.9} & {77.3}  & {73.5}  & {73.5} & {15.3} & {51.0} \\ 
            \bottomrule
        \end{tabular}
        }
    \label{tab:epoch-clevrer}
\end{table}

\subsection{Evaluating the Importance of Vision Module}
We illustrate the importance of the vision module in a VLM for physical reasoning by conducting the following experiment. Here, we QA-fine-tune only the language model part of PaliGemma-3B while freezing the vision module. The results in \cref{tab:vision} shows that the performance across all categories drops slightly for the language model-only setting. Therefore, jointly fine-tuning both the vision and language modules is essential for optimal performance, as it enables the model to better align visual features with linguistic representations.

\begin{table}[h!]
    \centering
    \caption{Performance drop due to freezing the vision module on the PaliGemma-3B-base model for the QA fine-tuning over CLEVRER.}
        {\small
        \begin{tabular}{ccccccc}
            \toprule
           {\textbf{Descriptive}} & \multicolumn{2}{c}{\textbf{Explanatory}} & \multicolumn{2}{c}{\textbf{Predictive}}& \multicolumn{2}{c}{\textbf{Counterfactual}}\\
            \cmidrule(lr){2-3} \cmidrule(lr){4-5} \cmidrule(lr){6-7} 
              & per ques. & per opt. & per ques. & per opt. & per ques. & per opt.\\
            \midrule       
	          -1.6 & -1.8 & -0.6  & -8.9  & -1.0 & -2.8 & -1.2 \\         
            \bottomrule
        \end{tabular}
        }
    \label{tab:vision}
\end{table}

\clearpage
\section{Falling Tower Dataset}
\label{appx:falling-tower}
The Falling Tower dataset is a benchmark for stability detection of stacked objects, inspired by the ShapeStacks benchmark~\cite{groth2018shapestacks}. It includes 4864 unique scenes, 72,775 questions, and detailed simulation-generated annotations to support training Vision-Language Models (VLMs) for spatial and physical reasoning. Each simulation instance is represented as a JSON file containing:
\begin{itemize}[leftmargin=*]
    \item \textbf{Scene Description:} A list of objects stacked from bottom to top with their respective offsets, e.g., ``Scene description: Here are the parts stacked from bottom to top: purple cube, yellow cylinder. Offsets for each part, from bottom to top, are: (-0.03, -0.05), (0.0, 0.02).''
    \item \textbf{Simulation Metadata:} Physical and rendering settings, including stability status (\texttt{stable: true/false}), the number of objects, gravity parameters, and camera settings.
    \item \textbf{Objects:} Detailed information about each object, including its type (e.g., cube, cylinder), dimensions, colors (both RGBA and HEX), rigid body properties (e.g., mass, friction), initial and final positions, and positional offsets. Rigid body properties used for simulation were fine-tuned to reflect real-world dynamics, enabling us to achieve 89\% accuracy in a human evaluation of 50 examples for stability detection.
    \item \textbf{Questions and Answers:} A variety of descriptive and stability QAs aimed at assessing spatial and physical reasoning, e.g:
    \begin{itemize}
        \item \textbf{Descriptive Questions:} ``How many objects are in the scene?'' (Answer: 2), ``What is the shape/color of the object above the purple cube?'' (Answer: yellow cylinder).
        \item \textbf{Stability Questions:} ``Will this collection of objects stay stationary?'' (Answer: False), ``Will the yellow cylinder stay stationary?'' (Answer: False).
    \end{itemize}
\end{itemize}

\begin{figure}[t]
    \centering
    \includegraphics[width=0.6\linewidth]{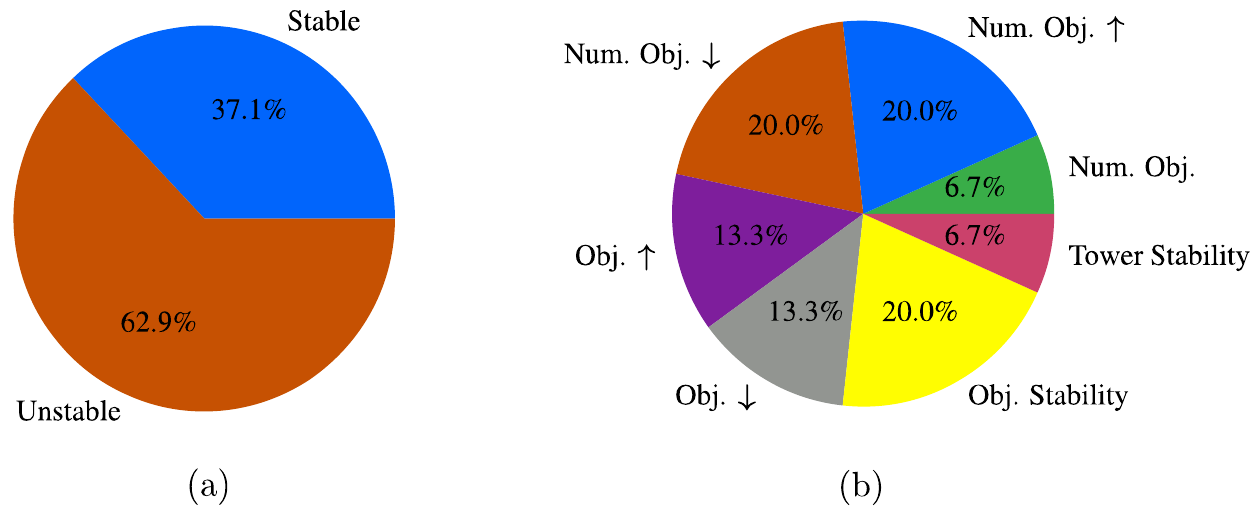}
    \caption{Falling Tower dataset: (a) distribution of stacked objects in terms of their stability, and (b) distribution of question types, including both descriptive and stability categories.}
    \label{fig:falling-tower-pie}
\end{figure}

\noindent \cref{fig:falling-tower-pie} shows the distribution of object stacks in terms of their stability, as well as the distribution of question types.

The Sim2Real dataset consists of 20 images. Seven stable and seven unstable cases were captured against a clean background, while six additional stable cases were captured with a varying background for testing the robustness of a vision model. Additionally, the dataset includes 100 human-generated questions, with five questions per image. The objects are 3D-printed using a J55™ Prime 3D Printer.\\

\noindent\textbf{Dataset Links:}
\begin{itemize}
    \item \href{https://drive.google.com/file/d/1RgpIdcywqvjYYXT3xTo-r1vxpUu2giby/view?usp=sharing}{Falling Tower Dataset}
    \item \href{https://drive.google.com/file/d/1Nd8s0Ik12Mfke5rHpkT0ZY7tOGIg9CYu/view?usp=sharing}{Sim2Real Dataset}
\end{itemize}
\clearpage
\section{Additional Experiments}
\subsection{Specialized Baselines for CLEVRER}
\label{appx:spec-clevrer}
Here, we compare the fine-tuned PaliGemma-3B model on the CLEVRER QA dataset to specialized architectures designed specifically for CLEVRER. Although the fine-tuned model does not outperform all benchmarks, its comparable performance highlights the potential benefits of generalist models over bespoke baselines.
\begin{table*}[ht]
\centering
\caption{Per-question performance of fine-tuned PaliGemma-3B compared to specialized methods on CLEVRER.}
\begin{adjustbox}{width=0.9\linewidth,center}
{
\begin{tabular}{llccccccc}
\toprule
\textbf{Category} & \textbf{Model} & {\textbf{Descriptive}} & {\textbf{Explanatory}} & {\textbf{Predictive}}& {\textbf{Counterfactual}}\\
\midrule
\multirow{4}{*}{\textbf{\shortstack{Specialized Methods}}} 
& VRDP \citep{ding2021dynamic} & 89.80 & 82.40 & 83.80 & 75.70 \\
& DCL \citep{chen2021grounding} & 90.70 & 82.80 & 82.00 & 46.50 \\
& CRCG \citep{ishay2024think} & 95.55 & 99.81 & 76.64 & 78.31 \\
& Aloe \citep{ding2021attention}& 94.00 & 96.00 & 87.50 & 75.60 \\
\midrule
\textbf{Fine-tuned QA} & {PaliGemma-3B-Fine-Tuned} & {92.90}& {94.70}  & {83.60} & {68.40} \\
\bottomrule
\end{tabular}
}
\end{adjustbox}
\label{tab:base-spec}
\end{table*}

\subsection{The Effect of PCBs on the InternVL 3.0 Model}
\label{appx:add-intern}
\begin{table*}[ht]
\centering
\caption{Performance of InternVL 3.0 (8B parameters), augmented with Physics Context Builders (PCBs), compared to its zero-shot version on the Falling Tower benchmark. HN refers to the Human Narration-style PCB. The second value after the slash indicates the Sim2Real accuracy, and the third value represents the F1 score on Sim2Real.}
\begin{adjustbox}{width=0.9\linewidth,center}
{\footnotesize
\begin{tabular}{lccccc}
\toprule
\textbf{Model} & \multicolumn{3}{c}{\textbf{Descriptive} [sim acc. / real acc.]} & \multicolumn{2}{c}{\textbf{Stability} [sim acc. / real acc. / real F1]} \\
\cmidrule(lr){2-4} \cmidrule(lr){5-6}
&num. obj. & num. obj. $\uparrow\downarrow$ & obj. $\uparrow\downarrow$ & obj. stable & tower stable \\
\midrule
InternVL3-8B & 81.57 / 78.95 & 52.77 / \textbf{78.95} & 53.85 / 84.21 & 52.71 / \textbf{84.21} / \textbf{80.19} & 46.42 / 73.68 / 68.64 \\
\midrule
InternVL3-8B-PCB & \textbf{95.94} / \textbf{88.24} & \textbf{66.29} / 70.58 &  \textbf{70.01} / \textbf{94.12} & \textbf{69.21} / 76.47 / 73.39 & \textbf{83.07} / \textbf{76.47} / \textbf{73.73}\\
\bottomrule
\end{tabular}
}
\end{adjustbox}
\label{tab:tower-internvl-pcb}
\end{table*}